\documentclass[a4paper,fleqn]{cas-dc}
\usepackage{float}
\usepackage[numbers]{natbib}
\usepackage[numbers]{natbib}
\usepackage{hyperref}
\usepackage{cleveref,multirow}
\usepackage{CJKutf8, multirow}
\usepackage{color,xcolor}
\usepackage{algorithm}
\usepackage{algpseudocode}
\usepackage{amsmath}
\hypersetup{
	colorlinks=true,
	urlcolor=cyan,
	linkcolor=purple,
	citecolor=blue}
\usepackage{soul}
\def\tsc#1{\csdef{#1}{\textsc{\lowercase{#1}}\xspace}}
\tsc{WGM}
\tsc{QE}
\begin{document}
\let\WriteBookmarks\relax
\def\floatpagepagefraction{1}
\def\textpagefraction{.001}

\shorttitle{EPAG: A Novel Enhanced Move Recognition Algorithm}    
\shortauthors{H.Wen et al.}

\title[mode = title]{EPAG: A Novel Enhanced Move Recognition Algorithm Based on Continuous Learning Mechanism with Positional Embedding }  

\author[1]{Hao \textcolor{black}{Wen}}[style=chinese,
        style=chinese,
       auid=000,
       bioid=1]
\fnmark[1] 
\ead{smczg@126.com} 
\credit{Conceptualization,Writing - Review & Editing,Supervision}

\author[1]{\textcolor{black}{Jie} Wang}[type=editor,orcid=0009-0007-6386-7655]
\cormark[1] 
\fnmark[2] 
\ead{zql02200059fs@163.com}
\credit{Conceptualization of this study, Methodology, Software, Writing - Original Draft,Visualization}

\author[2]{\textcolor{black}{Xiaodong} Qiao}
\fnmark[3] 
\ead{qiaox@wanfangdata.com.cn}
\credit{Data Curation,Project administration}

\address[1]{College of Information and Control Engineering, Xi'an University of Architecture and Technology,Shaanxi, China}
\address[2]{Wangfang Data, Beijing, China}

\cortext[1]{Corresponding author} 


\begin{abstract}
The identification of abstracts plays a vital role in efficiently locating the content and providing clarity to the article. Existing algorithms for move recognition exhibit a deficiency in their capacity to acquire word adjacent position information when word changes in Chinese expressions to obtain contextual semantics changes. This paper introduces EPAG: a novel enhanced move recognition algorithm with the improved pre-trained framework and downstream model for unstructured abstracts of Chinese scientific and technological papers. The proposed algorithm first performs data segmentation and vocabulary training. The EPAG framework is leveraged to incorporate word positional information, facilitating deep semantic learning and targeted feature extraction. Experimental results demonstrate that the proposed algorithm achieves 13.37$\%$ higher accuracy on the split dataset than on the original dataset and a 7.55$\%$ improvement in accuracy over the basic comparison model.
\end{abstract}

\begin{highlights}
\item This paper presents an enhanced EPAG framework for focused feature extraction, which enables deep semantic learning and improves the recognition performance of the model.
\item The enhanced EPAG framework combines a large-scale text corpus and a knowledge graph with a domain-specific corpus in the training of the sentencepiece model, which contributes to the model's contextual learning. Furthermore, the position information of each segment is embedded into the pre-trained framework to enhance the model's understanding of different semantic expressions. A continuous learning mechanism is used to account for long-distance dependencies and strengthen context representation.
\item Combining the handling of long and complex sentences with domain-specific knowledge training enhances the model's comprehension of nested sentence structures and domain-specific semantics.
\item The effectiveness of the proposed model is validated through experimental comparison with multiple baseline models.
\end{highlights}


\maketitle

\section{Introduction}
\label{1}

Text classification is the process of categorizing text content based on specific classification criteria \cite{Kowsari2019TextCA}. The classification of abstract structure and function is called move recognition and falls under the domain of text classification. The abstract is a concise summary of the entire article, providing an overview of the content \cite{article2}. The move of the abstract is to provide a concise summary of the section \cite{article3}, slicing language fragments with fixed semantics and originally proposed by Swales in 1981. Concise and clear words in an abstract is crucial for summarizing the main function of a sentence and helping readers quickly locate the content of an abstract. Text classification has been strongly associated with artificial intelligence recommendation systems, library and information science and information mining \cite{Kowsari2019TextCA}. It plays a significant role in other tasks such as entity recognition \cite{G2023120440,LE2023100015}, sentiment analysis \cite{AKHTER2023100027,ZHU2023126488}, paper retrieval, knowledge discovery, and knowledge graph construction \cite{DBLP:journals/corr/abs-2107-00931}, multimodal research \cite{RINALDI2023101107,DE2023100023}, et al.

Currently, public corpus of text classification mostly focuses on public opinion, news, Q$\&$A, disease, and topic analysis. However, it is  primarily based on English datasets. With the surge in the number of articles published in various regions, researchers' paper reading confronted with high time consuming, there is a lack of methods that can celerity retrieve abstract moves, especially for unstructured abstracts that do not clearly indicate move labels within a large paragraph of text. Too many text descriptions overload readers with information, making it difficult to quickly extract the key information. Therefore, there is an urgent need for abstract moving recognition to help researchers improve their paper reading efficiency and catch the key points of an article.

The advancement of deep learning has facilitated the application and development of neural networks in text classification applications. BERT (Bidirectional Encoder Representation from Transformers) has significantly improved the accuracy of various applications through the framework of pre-training and fine-tuning \cite{devlin2019BERT}. While the BERT model performs well on English corpora, it fails to consider the inter-character relationships within Chinese words and the relationships between masked characters and word boundaries.Text classification presents a challenging problem due to the high dimensionality of the text, especially in the case of scientific abstract in Chinese. These texts contain domain-specific terminology, which makes them difficult to understand. Furthermore, the complex sentence structures in the text present challenges in sequence recognition. The pre-trained model lacks sufficient knowledge in specific domains, leading to difficulties in accurately understanding domain-specific terminology and text semantics. Additionally, the model's inability to retain information from complex sentences affects the effectiveness of downstream sequence labeling tasks.

The contributions of this study are as follows$\colon$ 

1. Distinguishing and splitting techniques are employed to enhance model’s comprehension of long and complex sentences formed by nested sentence structures. Sentencepiece model \cite{kudo2018sentencepiece} is trained and utilized in this framework for boosting model’s semantic modeling capability in Chinese domain-specific scientific abstracts.

2. To address the issue of BERT model's single word boundaries and masked characters is not the most suitable framework for Chinese characters, this article incorporates the EPAG framework (EP-ERNIE$\_$AT-GRU). The pre-trained ERNIE \cite{liu2023ernie}, which improves the BERT model by considering the basic composition of Chinese language, introducing different masks for entities, phrase and large-scale knowledge graph for enhancing model's semantic modeling capability. Specifically, text position information is added and self-attention is modified to create the EP-ERNIE (Enhanced Positional ERNIE) model. Sentencepiece method  for word segmentation and centralized training that emphasizes domain-specific semantic learning in the EP-ERNIE model.

3. This study also considers contextual information, specifically the contribution of words in sentences to text classification, and uses the attention mechanism \cite{vaswani2023attention} to extract significant semantic information and emphasize feature recognition through AT-GRU.

This article is organized as follows: In Section~\ref{2}, we provide an overview of the studies that focus on the classification tasks. Section~\ref{3} introduces the details of the proposed EPAG framework. Section~\ref{4} explains the processing of the dataset in this study. Section~\ref{5} presents the experimental results and related discussion. Finally, Section~\ref{6} provides a conclusion of this study.

\section{Related work}
\label{2}
Traditional machine learning methods such as Support Vector Machine (SVM), Bayes, and Random Forest, have been used for text classification \cite{article12,Kowsari2019TextCA}. These methods showcase the efficiency brought by machine learning and promote research progress. However, these models have a strong dependence on features such as words, fixed collocation or computational rules, which limits their ability to deeply learn the semantics of the text. Therefore, the effectiveness of these methods is limited.

The innovative algorithms of traditional methods, such as the Logistic Regression Matching Pursuit (LRMP) \cite{LI2023110761} method and Topic modeling's application \cite{CHANDRAN2023100949} have demonstrated promising results and hold reference significance. The combination of the Lion Optimization Algorithm (LOA) and Neural Network Algorithm \cite{OROSOO2023100852} showcases the effectiveness of traditional algorithms when integrated with  new approaches.

In recent years, the performance of pre-trained Language Models (PLMs) in various natural language processing tasks has significantly improved. Researchers have optimized and transformed pre-trained models, leading to a surge in research activities. These models deeply learn the relationships between features and semantics of words in multilingual sentences through Transformer. Subsequent advancements, such as RoBERTa \cite{liu2019roBERTa} and ALBERT \cite{lan2020alBERT}, have also exhibited promising outcomes. The combination of deep learning models and pre-trained models in research \cite{JANSEN2023100020,Cui_2021,minaee2021deep,CHEN2023108751,KAUR2023108699} has demonstrated the potential application of pre-trained models in conjunction  with other deep learning fusion models in move recognition and other tasks such as automated translation \cite{YE2022100680}, online FAQs \cite{PEYTON2023100856}, and text migration applications in the field of engineering consulting \cite{JIANAN2023,OLIAEE2023100007}. Furthermore, the utilization of graph neural networks in NLP tasks has shown great potential and effectiveness. Existing graph-based semi-supervised short text classification methods and the Commonsense Knowledge-Powered Heterogeneous Graph Attention Network \cite{WU2023120800} have proven to be effective in text representation and classification.

For the research on language move recognition in Chinese scientific and technological abstracts, Ou \cite{XDQB202111001} conducted a study that combined BERT with traditional machine learning deep forest methods to classify English scientific papers in the field of chemistry based on seven language moves. Ding's research \cite{XDTQ201911002} introduced multiple deep learning methods such as DNN, LSTM, and Attention BiLSTM to recognize the structure and function of scientific paper abstracts. The study verified that the BiLSTM model, which incorporates the Attention mechanism, achieved the best results. Wang \cite{XDTQ202006008} improved the BERT input layer and incorporated the relative position of each sentence in the entire abstract and identified 11 step labels in the main text of academic papers. These enhancements resulted in good recognition results. Liu \cite{XDTQ202308008} proposed a method to enhance the learning of abstracts in social science academic texts by incorporating domain data in each stage of model pre-training, fine-tuning, and model output. Du's research \cite{XDTQ20230223001} constructs a move classification dataset for academic papers using multi-stage fine-tuning. They also propose a move recognition method based on the BERT model, which incorporates the section title text to recognize moves at a fine-grained level. Mao \cite{XDTQ20230814001} combined SCI-BERT-BiLSTM-CRF with Active Learning for Tagging Strategy in the recognition of medical literature abstract moves.

The attention mechanism was first introduced in the field of image processing \cite{article18}, which mimics human vision by focusing on the key points while reducing attention to other information. Attention mechanism has a significant impact on natural language processing \cite{article19,article20}. Particularly which can focus more on the key points of the text, improve network forgetting, and simplify the network for deep learning of multi-layer superposition efficiently, optimizing the output results for long text inputs. Studies \cite{ref21,perez2022topological} have demonstrated the excellent performance of the fusion model for attention mechanism and deep learning network in text classification. Therefore, this article utilizes AT-GRU for feature extraction.

Based on the above analysis, this article utilizes the EP-ERNIE model to address issues with word segmentation and mask adaptation that BERT encounters. The EP-ERNIE model is more suitable for Chinese text processing. Additionally, domain-specific knowledge and text position information is added to the model, self-attention for EPGA framework is modified and long-complex sentences are handled to conduct research on the automatic recognition of language moves in Chinese scientific and technological abstracts.

\section{Model}
\label{3}
\begin{figure}[ht]
	\begin{center}
		\includegraphics[width=10.0cm]{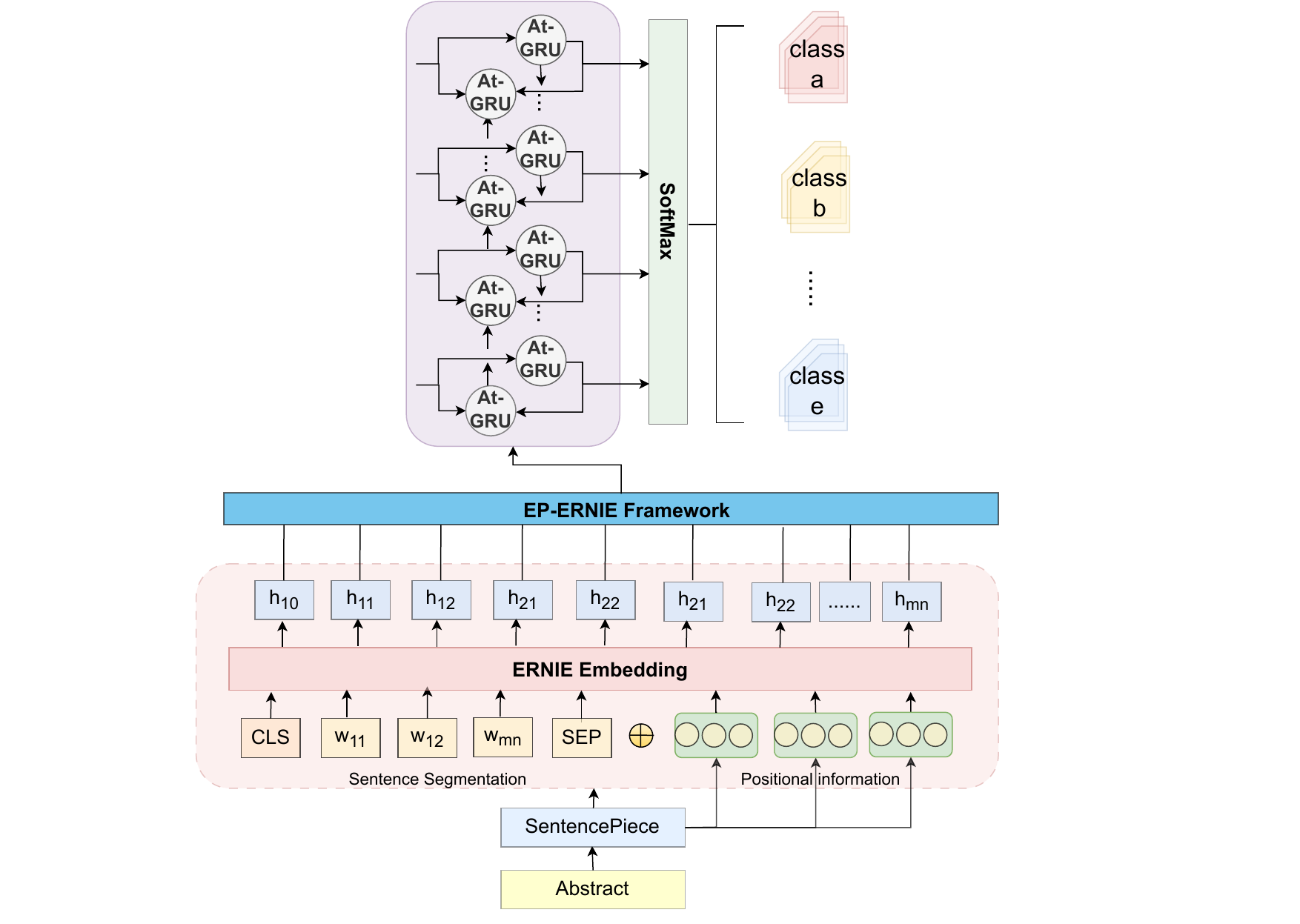}
		\caption{EPAG Structure}
		\label{fig1}
	\end{center}
\end{figure}

This article proposes an approach based on the EPAG: (EP-ERNIE$\_$AT-GRU) algorithm for recognizing Chinese abstract moves. The algorithm consists of EP-ERNIE for deep semantic learning and AT-GRU for feature recognition. The model structure is illustrated in Figure \ref{fig1}.

The given abstract contains N sentences, $Abstract$ = [{${s_{1},s_{2},\dots,s_{N}}$}]. Sentencepiece model is trained to partition sentences into sequences and perform position encoding for deep semantic learning. EP-ERNIE embedding is used to vectorize the text into 768 dimensions for model training and obtaining the hidden word vector. EP-ERNIE's framework is employed for contextual understanding, obtaining vector features, and learning the internal connections not only through phrase entity levels, but also word positional information. AT-GRU uses a bidirectional gated network with an attention mechanism for targeted feature recognition. Finally, the recognized move are obtained through the softmax layer to accomplish automatic move recognition.

\subsection{EP-ERNIE Deep Semantic Learning}

Firstly, the EP-ERNIE pre-trained model is combined with the trained Sentencepiece model to perform domain-specific deep semantic learning. By leveraging the model's pre-trained advantages in large-scale prior knowledge like entities and words, contextual understanding, feature learning, and expertise knowledge, the model focuses on learning the internal connections of abstract concepts. It integrates the learned knowledge with phrase and entity level mask information to express fused text semantic word vectors.  Through training, the model obtains vector features of text data.

\subsubsection{ERNIE Pre-trained Model}	

\begin{figure}[ht]
	\centering
	\begin{minipage}[t]{0.48\textwidth}
		\centering
		\includegraphics[width=9.0cm]{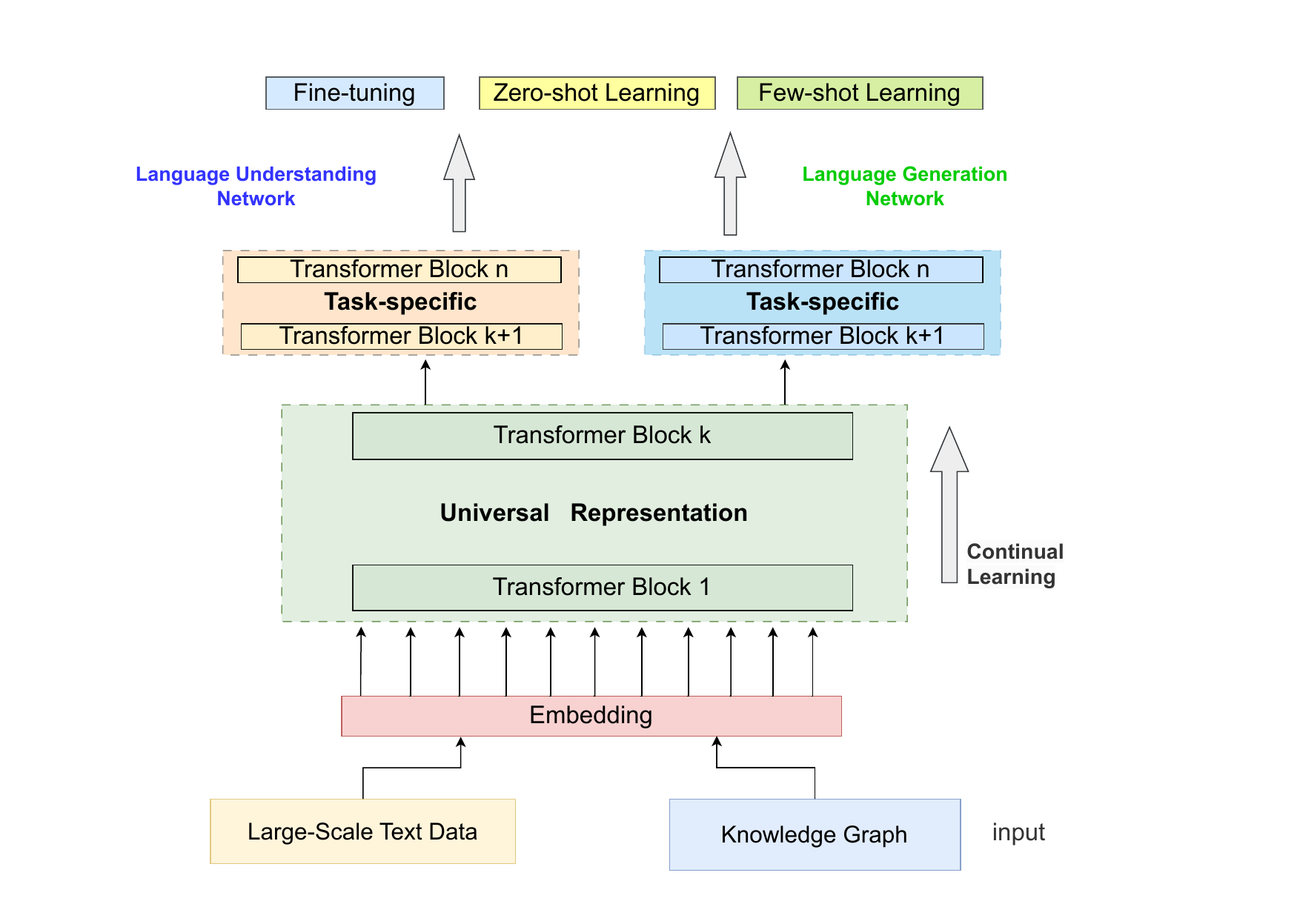}
		\caption{ERNIE Structure.}
		\label{fig2}
	\end{minipage}
	\begin{minipage}[t]{0.48\textwidth}
		\centering
		\includegraphics[width=9.0cm]{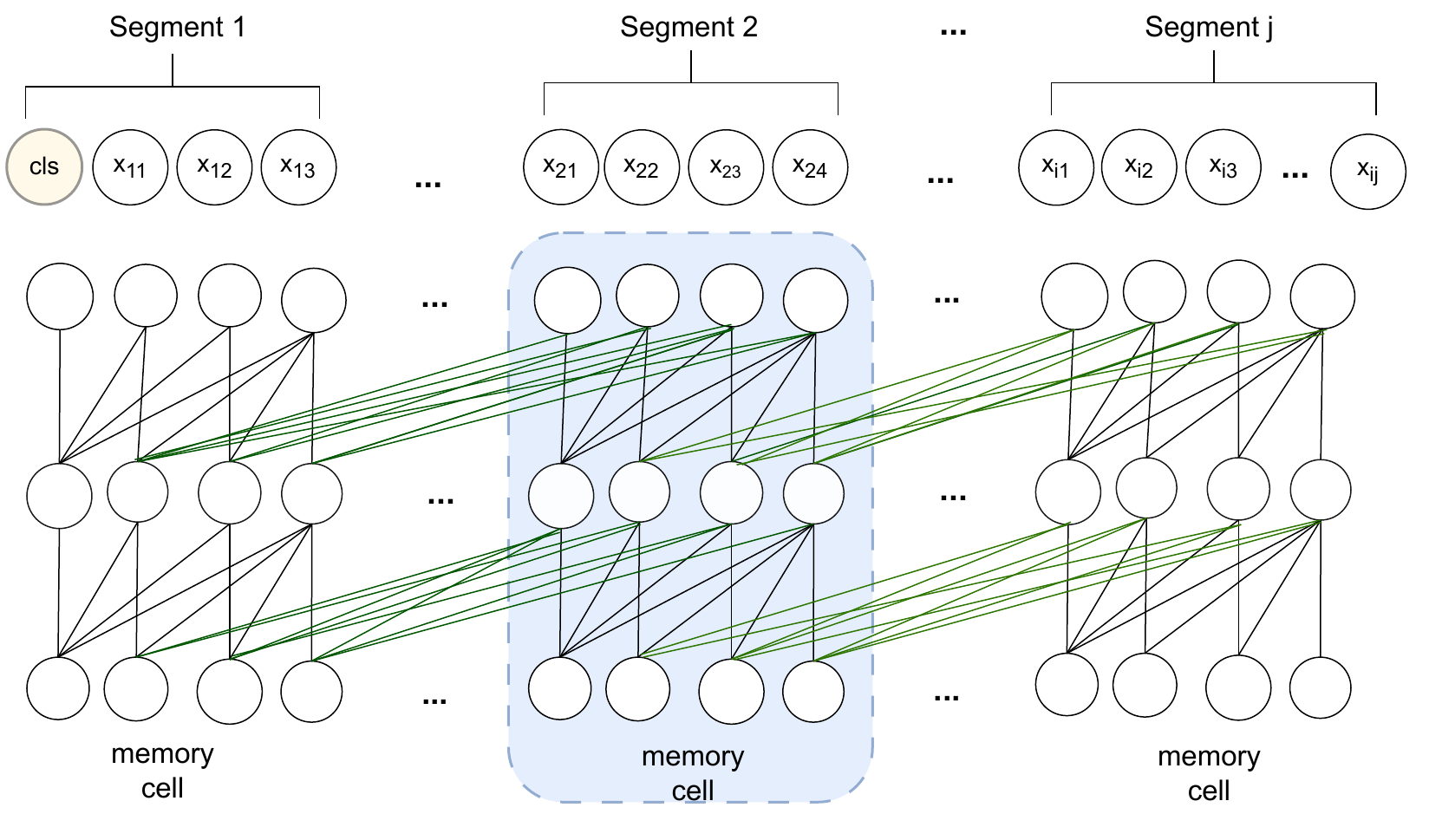}
		\caption{Transformer-XL’s Fragment Regression.}
		\label{fig3}
	\end{minipage}
\end{figure}

ERNIE is a universal multi-mode pre-trained framework for large-scale knowledge reinforcement, prior knowledge graph fusion, and continuous learning in language understanding and generation.  The structure of ERNIE is shown in Figure \ref{fig2}.

ERNIE is pre-trained with large-scale text and knowledge graphs. Its underlying network structure: Universal Representation employs Continuous Learning to improve its ability to capture lexical and syntactic language features. The task-specific layer utilizes two types of network structures that focus on natural language understanding, generation tasks, and specific feature extraction. In order to describe complex textual semantic relationships, ERNIE's mask strategy is divided into different granularity, word-aware pre-training task for organizing lexical relationships between words, structure-aware task for ordering the syntactic relationship between sentences and semantic-aware for semantic relationships between semantic blocks of organizational text. It can be seen, ERNIE's mask method is designed for Chinese characters and is more suitable for semantic learning of Chinese data, and with abundant input knowledge. This article uses ERNIE as the primary pre-trained model.

\subsubsection{Continuous Learning Mechanism}		
This pre-training model's deep learning mainly relies on Transformer-XL's long-distance calculation \cite{dai2019transformerxl}. The fragment regression mechanism in the Transformer is utilized to introduce a memory unit that caches the previous fragment's hidden state and updates the next fragment's hidden state. This iterative process models the connections between fragments, enabling continuous learning to be achieved. Figure \ref{fig3} illustrates how Transformer-XL embeds words into vectors using memory units.  It also demonstrates how the model links context during the training process and preserves the hidden state of long-distance dependencies. This continuous learning approach provides a longer effective context, resolves context fragmentation, and enhances the text's semantic representation.

\subsubsection{Improved Self-Attention}
Self-attention in Transformer-XL is a process that involves linear mapping and weighted summation of input word vectors \cite{ref24}. The calculation results remain the same when the text order is changed, indicating that the text order is not closely related to the key points displayed by the text semantics. However, in reality, changing the word order can significantly alter sentence semantics, as seen in expressions "\begin{CJK*}{UTF8}{gbsn}屡败屡战\end{CJK*} (comeback attempts-repeatedly)" and "\begin{CJK*}{UTF8}{gbsn}屡战屡败\end{CJK*} (have suffered repeated defeats)", "\begin{CJK*}{UTF8}{gbsn}科学\end{CJK*} (science)" and "\begin{CJK*}{UTF8}{gbsn}学科\end{CJK*} (subject)". Similarly, the positional information of words in a sentence has a significant impact on semantics, as demonstrated by sentences like "\begin{CJK*}{UTF8}{gbsn}我到了\end{CJK*} (I'm here)" and "\begin{CJK*}{UTF8}{gbsn}到我了\end{CJK*} (It's my turn)". Due to the differences in language structure, Chinese characters are formed by words, whereas Indo-European languages like English use characters as the minimum unit. This distinction contributes to the uniqueness of this issue. In previous research, modeling word position information in text involved adding position encoding and incorporating position information into model training \cite{zhou2023sat,yun2023dynamic}. However, this method requires modifying the architecture of the original pre-trained model and performing additional calculations, which is not suitable for creating a lightweight model.

This article addresses the impact of changes in sentence position information on sentence semantics by encoding segmented phrases and recording their positional information. This approach prevents changes in phrase order from affecting text semantics. 

The improvement in the model's position encoding draws on the optimization thinking of the attention mechanism in image processing \cite{chu2023conditional}. It internalizes position features into the self-attention calculation of the model,
\begin{equation}
    \label{eq1}
    \begin{split}
        x_{j} &= W_{e}w_{_i,_t},t\in[1,N] 
    \end{split}
\end{equation}

The word vector sequence $x_{j}$ has been encoded, $W_{e}$ is the weight parameter of the embedding layer for the pre-trained model. The embedding vector reflects the semantic relationship represented by the word vector sequence, then $x_{j}$ is input into the self-attention for calculation. To address the problem of the same text sequence producing different self-attention calculation results when the text order changes, the formula for the self-attention mechanism is modified as follows $\colon$
\begin{equation}
    \label{eq2}
    z_{i} = \sum_{j=1}^{n}a_{_i,_j}(X_{j}W^{v}+a_{_i,_j}^{v})
\end{equation}
\begin{equation}
	\label{eq3}
	a_{_i,_j} = \dfrac{\exp e_{_i,_j}}{\sum_{k=1}^{n}\exp e_{_i,_k}}
\end{equation}
\begin{equation}
	\label{eq4}
	e_{_i,_j} = \dfrac{x_{i}W^{Q}(x_{j}W^{k}+a_{_i,_j}^{k})^{T}}{\sqrt{d_{z}}}
\end{equation}

For input text features represented by $a_{i,j}^{k}$ and $a_{i,j}^{v}$, when the input position information $i$, $j$ are fixed, the text has a specific position and calculated weight information. The position information is derived from the data sequence after Sentencepiece’s segmentation. This modification effectively avoids the problem of the different text sequence producing same self-attention values, and ensure that the emphasis displayed by text semantics can be distinguished to enhance the model’s understanding of different semantic expressions.	

\subsection{AT-GRU Feature Recognition}
In the second part, the trained text vector features are input into the AT-GRU layer for 512 dimension. Attention mechanism calculates the weight information of different words in the text, strengthens the updated context features at different time periods, and the BiGRU network for feature recognition. The Softmax layer is then used for 5 labels classification.

The sentence encoding representation of implicit states before and after gating networks for the i-th sentence in the abstract is$\colon$

\begin{equation}
	\label{eq5}
	\overrightarrow{h_{i}} = \overrightarrow{GRU}_{(sen_{i})}
\end{equation}	
\begin{equation}
	\label{eq6}
	\overleftarrow{h_{i}} = \overleftarrow{GRU}_{(sen_{i})}
\end{equation}	

Sentence encoding representation of implicit states before and after gating networks$\colon$

\begin{equation}
	\label{eq7}
	h_{i} = \overrightarrow{[h_{i}},\overleftarrow{h_{i}]}
\end{equation}		

To effectively extract semantic information from long text and focus on information that is more suitable for classification into tags, this study aims to extract the hidden representation of words that plays a crucial role in recognizing moves in the text sequence.

\begin{equation}
	\label{eq8}
	u_{i} = \tanh(W_{s}h_{i}+b_{s})
\end{equation}		

An attention mechanism is added to calculate word weights$\colon$

\begin{equation}
	\label{eq9}
	a_{i} = \dfrac{\exp (u_{i}^{T})u_{s}}{\sum_{i}\exp (u_{i}^{T})u_{s}}
\end{equation}

By focusing on the semantic representation of key clause hidden information through an attention mechanism, weighted summation is used to obtain sentence information:

\begin{equation}
	\label{eq10}
	sen = \sum_{i} a_{i}h_{i}    
\end{equation}

A softmax layer calculates the probabilities of the text belonging to each category, and the category with the highest probability is considered as the classification result for the abstract.

\section{Data Processing}
\label{4}
\subsection{Datasets}

To enhance the training of the proposed model and address the urgent need for an effective method of abstract reading in move recognition, we utilized datasets extracted from unstructured papers in the Chinese scientific and technological field. A total of 10,200 complete unstructured Chinese scientific and technological paper abstract sentences were collected from Acta Electronica Sinica for this study.
 
\begin{figure}[ht]
	\begin{center}
		\includegraphics[width=9.0cm]{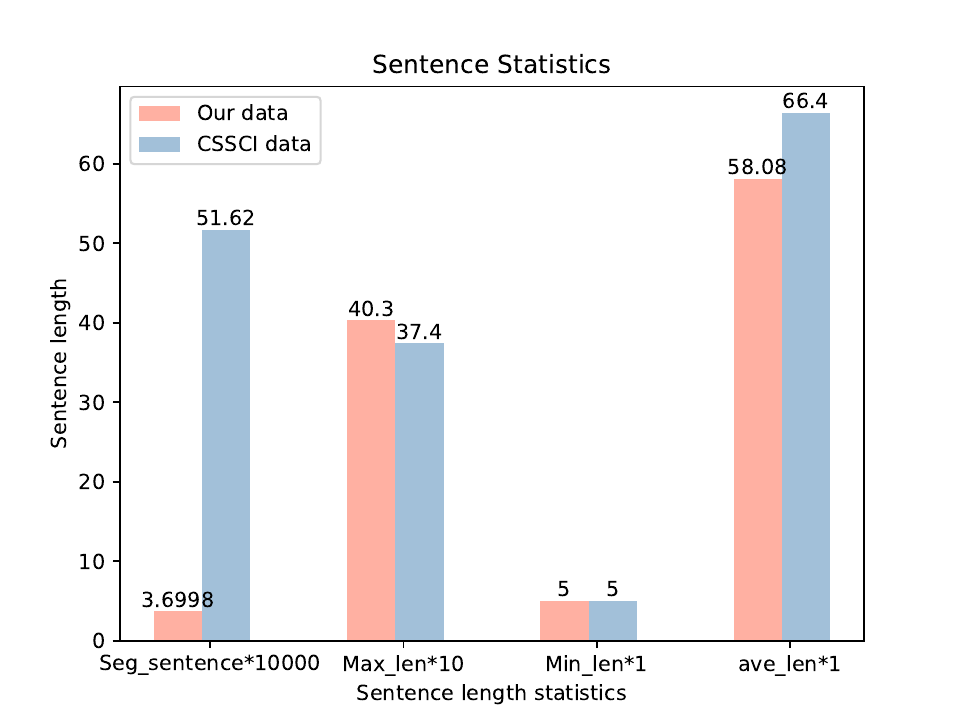}
		\caption{Data Distribution}
		\label{fig10}
	\end{center}
\end{figure}

\begin{figure}[ht]
	\begin{center}
		\includegraphics[width=7.0cm]{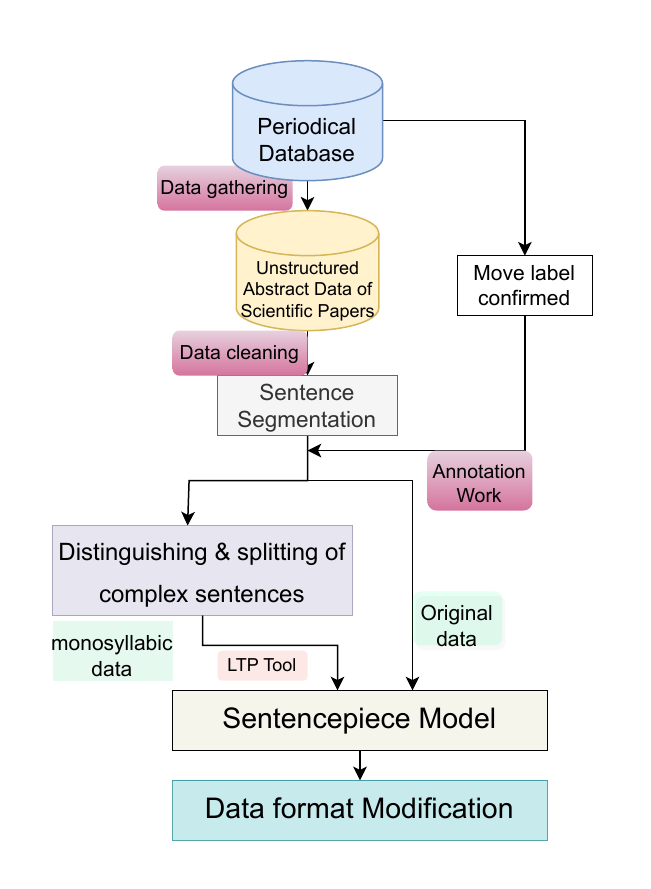}
		\caption{Data preprocessing procedure}
		\label{fig11}
	\end{center}
\end{figure}

The data preprocessing procedure, which encompasses multiple steps, is illustrated in Figure \ref{fig11}. Firstly, the acquired data underwent a cleaning process to eliminate extraneous characters, including spaces, tabs, and line breaks in order to obtain a well-structured abstract dataset. Next, the process of sentence segmentation was conducted, and the subsequent analysis involved recording the quantity and length of sentence units. Through a comprehensive review of relevant literature and meticulous analysis of unstructured abstract data, the various categories of move labels present in abstract data were determined. A comprehensive summary of all label types was generated, and suitable label terms that represent the move labels in Chinese scientific and technological paper abstract data were chosen for the purpose of this study. The sentences underwent a scientific and rational labeling process through annotation and comparison by language experts with relevant expertise. This process was conducted prior to the distinguishing and splitting of complex sentences, which formed the experimental data for training the sentencepiece model. The distribution of the data is illustrated in Figure \ref{fig10}. After completing the segmentation process, a total of 36,998 sentences were obtained. The longest sentence contains 403 words, whereas the average sentence length is calculated to be 58.08 words. To provide a comparative analysis, the CSSCI public abstract dataset comprises 516,200 sentences, with the longest sentence containing 374 words and an average sentence length of 66.4 words. In order to efficiently load the text into the model's iterator for training , we have included the vectorized step labels after the text. These labels are separated from the text using tab characters, which allows for automatic move recognition.

\subsection{Determination and tagging of move types}
Based on the results of literature survey and analysis of abstract structures, academic abstracts primarily focused on purpose, method, conclusion, and occasionally included descriptions of background, product, and other aspects. In order to enable the model to accurately identify moves in the abstract, this study categorizes moves into five labels: background, purpose, method, result, and conclusion.

The abstracts were labeled according to the identified move types. After several rounds of labeling and comparison by language experts, a consensus was reached, resulting in 10,200 sentences for the experiment.

\subsection{SentencePiece and LTP tools}
To enhance the model's comprehension of nested sentence structures and domain-specific semantics, the SentencePiece model and LTP tools were utilized in this experiment.

To improve the training of the vocabulary for Chinese scientific and technological abstracts, addressing the pre-trained model’s inadequacy in adapting to Chinese word segmentation and enhancing the model's semantic learning in the professional domain, this study incorporates the Sentencepiece model, which is an open-source tool developed by Google to address word segmentation in neural networks and is available at \textcolor{blue}{https://github.com/goole/sentencepiece}.  

\begin{figure*}[ht]
	\begin{center}
		\includegraphics[width=18.0cm]{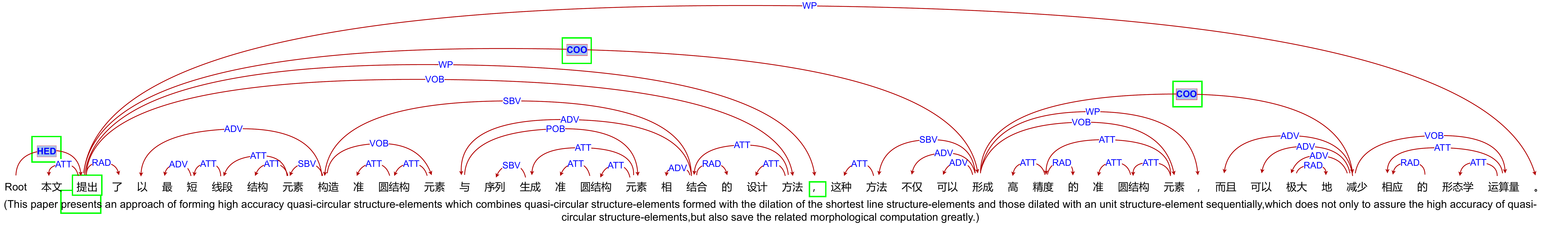}%
		\caption{Dependency Parser of LTP}
		\label{fig4}
	\end{center}
\end{figure*}

For text dependency analysis, the LTP (Language Platform Cloud) \cite{Che2010LTPAC} is employed, which provides 14 kinds of different dependency labels, such as SBV (subject-verb), VOB (verb-object), ATT (attribute), HED (the core word of a sentence) and is available at \textcolor{blue}{https://github.com/HIT-SCIR/ltp}. Drawing inspiration from research \cite{ref28}, our study uses the dependency direction between COO and HED dependencies obtained from LTP's syntax analysis results as an indicator for distinguishing and splitting complex sentences. Different dependency tags  indicate the semantic associations between language units. The starting word of the arrow is referred to as the parent node, while the word pointed by the arrow is considered the child node, as shown in Figure \ref{fig4}. The COO tag signifies that the language units at both ends of the arrow have a coordinate relationship, representing language units at the same semantic level within a sentence. Distinguishing and splitting complex sentences relies on dependency parsing, if the parent node of COO is HED, the sentence is segmented at the comma before COO to obtain the split single semantic data.

The algorithm for distinguishing and splitting complex sentences as \ref{alg1}:

\begin{algorithm*}[ht]
\label{alg1}
	\caption{Distinguishing and splitting complex sentences} 
	\hspace*{0.02in} {\bf Input:} 
	Abstract sentences: $Abstract$ = [${s_{1},s_{2},\dots,s_{N}}$]\\
	\hspace*{0.02in} {\bf Output:} 
	Splitted single semantic abstract sentences: $Monosyllabic$ = [${s_{1},s_{2},\dots,s_{N}}$]
	\begin{algorithmic}[1]
		\State Input abstract sentences 
		\For{ sentence in Abstract} 
		\State Split sentence with period ---$>$  $Sentence$ = [${c_{1},c_{2},\dots,c_{N}}$]
		\State LTP dependency parsing ( as Figure \ref{fig4} ) ---$>$  $Dependency$ = [${'seg':'t_{1}\dots t_{N}', 'id':'1\dots N'}$, ${'Parent\_id':'1\dots N'}$, ${ 'Dependency\_tag':'HED, WP, ATT, POB, COO ..'}$]
			\For{ t in id} 
			\If{ COO's Parent$\_$id == HED's id}
			\State split the comma before COO ---$>$   $Sentence$ = [${splited_{t1}, splited_{t2}\dots}$]
			\State Monosyllabic.append(${splited_{t1}, splited_{t2}\dots}$)
			\Else   $Monosyllabic$ = {${Sentence}$}
			\State Monosyllabic.append(${Sentence}$)
			\EndIf
			\EndFor
		\EndFor  \\
	\Return Monosyllabic
	\end{algorithmic}
\end{algorithm*}

An example of splitting complex sentences is as Figure \ref{fig12}:
\begin{figure}[ht]
	\begin{center}
		\includegraphics[width=7.0cm]{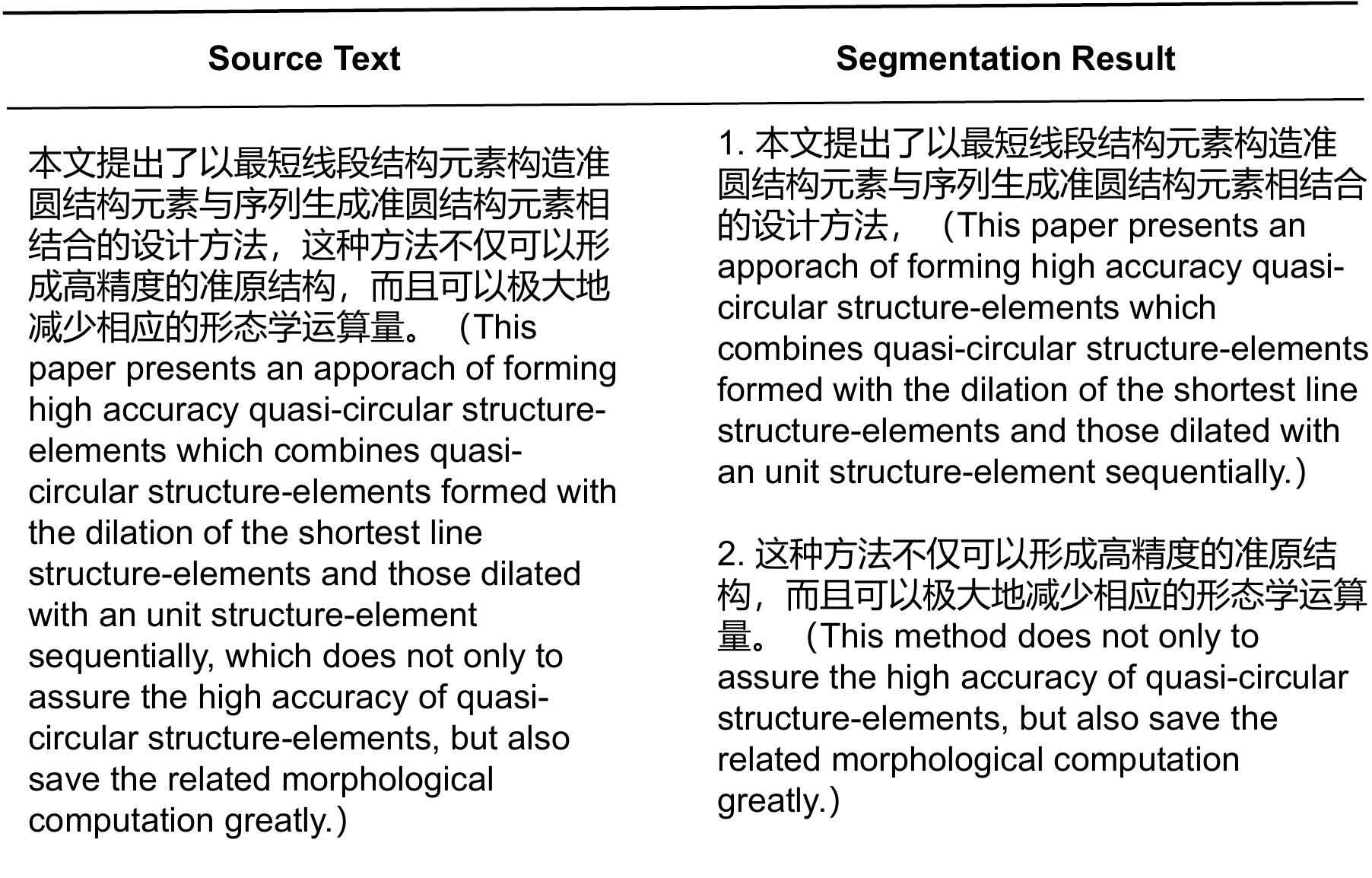}
		\caption{An example of splitting complex sentences}
		\label{fig12}
	\end{center}
\end{figure}

\subsection{Overall process}
\begin{figure}[ht]
	\begin{center}
		\includegraphics[width=7.0cm]{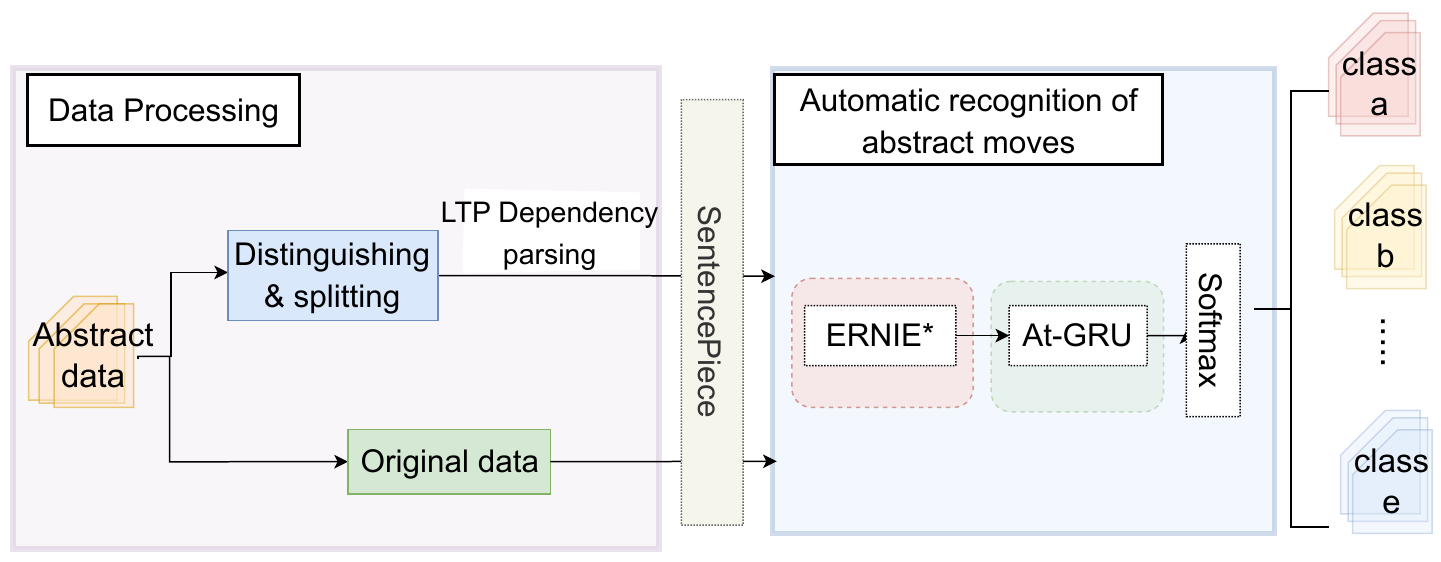}
		\caption{Overall flowchart}
		\label{fig5}
	\end{center}
\end{figure}

The overall flowchart of the automatic recognition model is shown in Figure \ref{fig5}$\colon$

In the flowchart, LTP is used to analyze the dependency structure of an abstract corpus in order to identify and split complex sentences. The SentencePiece model is trained on the dataset to perform semantic learning and segmentation training for move recognition in EP-ERNIE$\_$AT-GRU. An AT-GRU structure is used for feature recognition. Finally, labels for automatic recognition of move steps are obtained.

\section{Experimental Setup}
\label{5}

\subsection{Experimental Datasets}
The original labeled data and the split single semantic data (monosyllabic data) mentioned in Section~\ref{4} consists of 10,000 pieces each. They are divided into training and test samples in an 8:2 ratio. The distribution of data labels is shown in figure {\ref{fig6}}:

\begin{figure}[ht]
	\begin{center}
		\includegraphics[width=8.0cm]{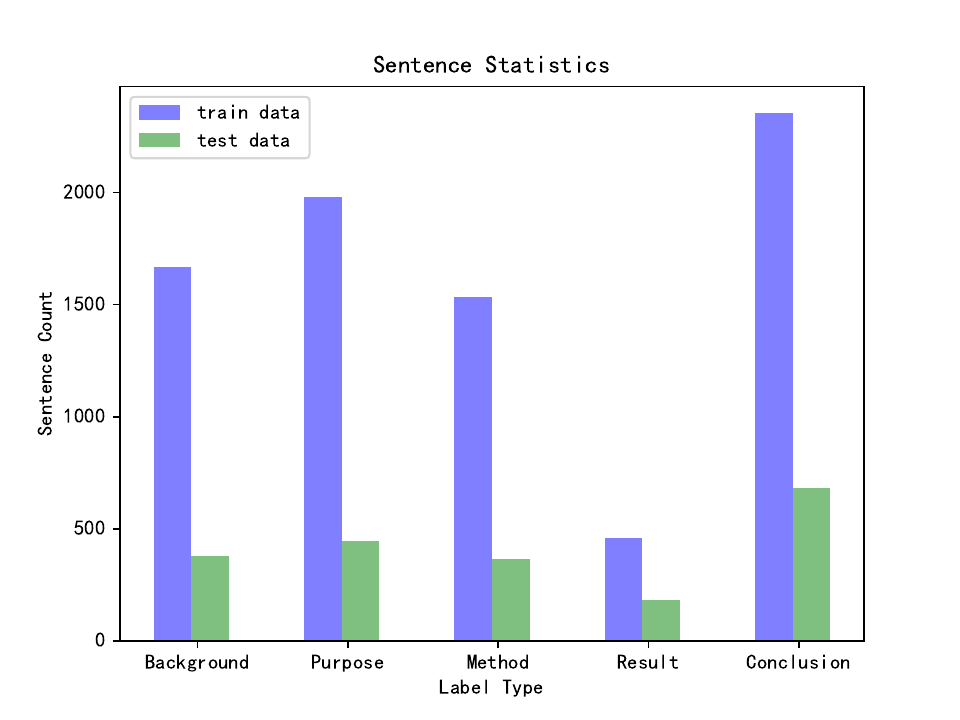}
		\caption{The distribution of data labels}
		\label{fig6}
	\end{center}
\end{figure}

\subsection{Experimental Settings}
The experiment contains three comparative content aspects. Experiment (1) compares the recognition effects of different datasets, the original dataset and the dataset with single semantic data. This comparison aims to determine the impact of complex sentence splitting on model recognition.

Experiment (2) analyzes the influence of different models on the accuracy of move recognition and verifies the efficiency of our model. 

Experiment (3) analyzes the influence of EP-ERNIE model compared with ERNIE and BERT on the accuracy of move recognition and verifies the effectiveness of the modified model. The comparison models are as follows:

(1) ERNIE/BERT model learns the semantic relationship of text, trains to obtain a 768-dimensional vector, and outputs the classification result through the fully connected layer.

(2) ERNIE-BiLSTM/BERT-BiLSTM model performs deep semantic learning through the ERNIE/BERT pre-training model, and the bidirectional long short-term memory network, which is classified by softmax layer.

(3) ERNIE-BiGRU/BERT-BiGRU model performs deep semantic learning through the ERNIE/BERT pre-training model, and the bidirectional gating network for move recognition.

(4) ERNIE-CNNGRU/BERT-CNNGRU model performs deep semantic learning through the ERNIE/BERT pre-training model and the gated network with convolutional network feature extraction to expand move recognition.

(5) ERNIE-textRcnn/BERT-textRcnn model inputs the obtained word vectors into the bidirectional loop structure (RNN) + pooling layer to extract context information for classification.

\subsection{Experimental Results and Analysis}
The experiment evaluates the proposed EPAG (EP-ERNIE$\_$AT-GRU) framework for automatic move recognition of scientific abstracts in both monosyllabic and original data for Chinese. 
Table \ref{tab1} and Table \ref{tab2} display the evaluation results. The best performance of each evaluation function on the Monosyllabic/Original data is represented in bold font.

\begin{table*}[ht]
	\caption{The evaluation results on EP-ERNIE and its comparative model.}
	\begin{center}
		\begin{tabular}{|c|@{\extracolsep{1em}}c|c|c|c|}
			\hline {Model} & {Data} & {Macro$\_$F1/$\%$} & {Precision/$\%$} & {Recall/$\%$}  \\
			\hline
			\multirow{2}{*}{EP-ERNIE} 
			& {Monosyllabic} & {87.0} & {89.55} & {86.33}  \\
			\cline{2-5}
			& {Original} & {72.8} & {77.35} & {77.31} \\
			\hline
			
			\multirow{2}{*}{EP-ERNIE-BiLSTM} 
			& {Monosyllabic} & {90.7} & {92.6} & {88.7}  \\
			\cline{2-5}
			& {Original} & {74.6} & {78.26} & {77.93} \\
			\hline
			
			\multirow{2}{*}{EP-ERNIE-BiGRU} 
			& {Monosyllabic} & {91.06} & {93.45} & {90.07} \\
			\cline{2-5}
			& {Original} & {75.08} & {78.95} & {76.38}  \\
			\hline
			
			\multirow{2}{*}{EP-ERNIE-CNNGRU} 
			& {Monosyllabic} & {\textbf{95.2}} & {96.9} & {95.3}  \\
			\cline{2-5}
			& {Original} & {76.82} & {\textbf{84.12}} & {\textbf{83.64}}  \\
			\hline
			
			\multirow{2}{*}{EP-ERNIE-textRcnn} 
			& {Monosyllabic} & {94.2} & {94.95} & {92.2}   \\
			\cline{2-5}
			& {Original} & {72.97} & {75.88} & {74.96}   \\
			\hline
			\multirow{2}{*}{(Our’s)EPAG} 
			& {Monosyllabic}  & {95.1} & {\textbf{97.1}} & {\textbf{95.4}}  \\
			\cline{2-5}
			& {Original} & {\textbf{77.09}} & {83.73} & {83.39}  \\
			\hline
		\end{tabular}
		\label{tab1}
	\end{center}
\end{table*}
 
\begin{table*}[ht]
	\caption{The evaluation results on ERNIE and its comparative model.}
	\begin{center}
		\begin{tabular}{|c|@{\extracolsep{1em}}c|c|c|c|}
			\hline {Model} & {Data} & {Macro$\_$F1/$\%$} & {Precision/$\%$} & {Recall/$\%$}  \\
			\hline
                \multirow{2}{*}{word2vec-BiLSTM}
			& {Monosyllabic} & {77.16} & {77.95} & {76.38}   \\
			\cline{2-5}
			& {Original} & {65.4} & {66.46} & {65.32}    \\
			\hline
                \multirow{2}{*}{word2vec-BiGRU}
			& {Monosyllabic} & {77.50} & {78.61} & {76.42}   \\
			\cline{2-5}
			& {Original} & {66.23} & {67.96} & {67.28}    \\
			\hline
                \multirow{2}{*}{BERT}
			& {Monosyllabic} & {80.91} & {81.58} & {81.30}   \\
			\cline{2-5}
			& {Original} & {70.33} & {71.53} & {71.31}    \\
			\hline
			    \multirow{2}{*}{ERNIE} 
			& {Monosyllabic} & {85.02} & {87.12} & {83.64}   \\
			\cline{2-5}
			& {Original} & {71.98} & {75.35} & {65.4}    \\
			\hline
			\multirow{2}{*}{BERT-BiLSTM}
			& {Monosyllabic} & {81.53} & {82.44} & {82.00}   \\
			\cline{2-5}
			& {Original} & {70.09} & {71.26} & {70.93}    \\
			\hline 
			\multirow{2}{*}{ERNIE-BiLSTM} 
			& {Monosyllabic} & {87.7} & {89.13} & {88.33}   \\
			\cline{2-5}
			& {Original} & {72.56} & {76.1} & {64.6}   \\
			\hline
			\multirow{2}{*}{BERT-BiGRU}
			& {Monosyllabic} & {83.89} & {84.22} & {83.64}   \\
			\cline{2-5}
			& {Original} & {72.87} & {73.47} & {73.26}    \\
			\hline
			\multirow{2}{*}{ERNIE-BiGRU} 
			& {Monosyllabic} & {89.06} & {90.36} & {89.92}   \\
			\cline{2-5}
			& {Original} & {72.96} & {76.7} & {\textbf{76.1}}   \\
			\hline
			\multirow{2}{*}{BERT-CNNGRU}
			& {Monosyllabic} & {89.89} & {90.75} & {90.18}   \\
			\cline{2-5}
			& {Original} & {73.98} & {75.3} & {74.98} \\
			\hline
			\multirow{2}{*}{ERNIE-CNNGRU} 
			& {Monosyllabic} & {91.52} & {93.71} & {92.47}   \\
			\cline{2-5}
			& {Original} & {\textbf{76.72}} & {\textbf{82.35}} & {74.6}   \\
			\hline
			\multirow{2}{*}{BERT-textRcnn}
			& {Monosyllabic} & {83.82} & {85.71} & {84.25}   \\
			\cline{2-5}
			& {Original} & {70.06} & {71.56} & {70.62}   \\
			\hline
                \multirow{2}{*}{ERNIE-textRcnn} 
			& {Monosyllabic} & {88.28} & {89.45} & {91.96}   \\
			\cline{2-5}
			& {Original} & {71.57} & {73.95} & {71.19}   \\
			\hline
			\multirow{2}{*}{BERT$\_$AT-GRU}
			& {Monosyllabic}  & {90.45} & {91.9} & {90.23} \\
			\cline{2-5}
			& {Original} & {74.55} & {76.7} & {71.4}   \\
			\hline
			\multirow{2}{*}{ERNIE$\_$AT-GRU} 
			& {Monosyllabic}  & {\textbf{91.58}} & {\textbf{94.17}} & {\textbf{93.35}} \\
			\cline{2-5}
			& {Original} & {76.59} & {82.0} & {73.4}   \\
			\hline
		\end{tabular}
		\label{tab2}
	\end{center}
\end{table*}

\begin{figure}[ht]
	\centering
	\begin{minipage}[t]{0.48\textwidth}
		\centering
		\includegraphics[width=8.0cm]{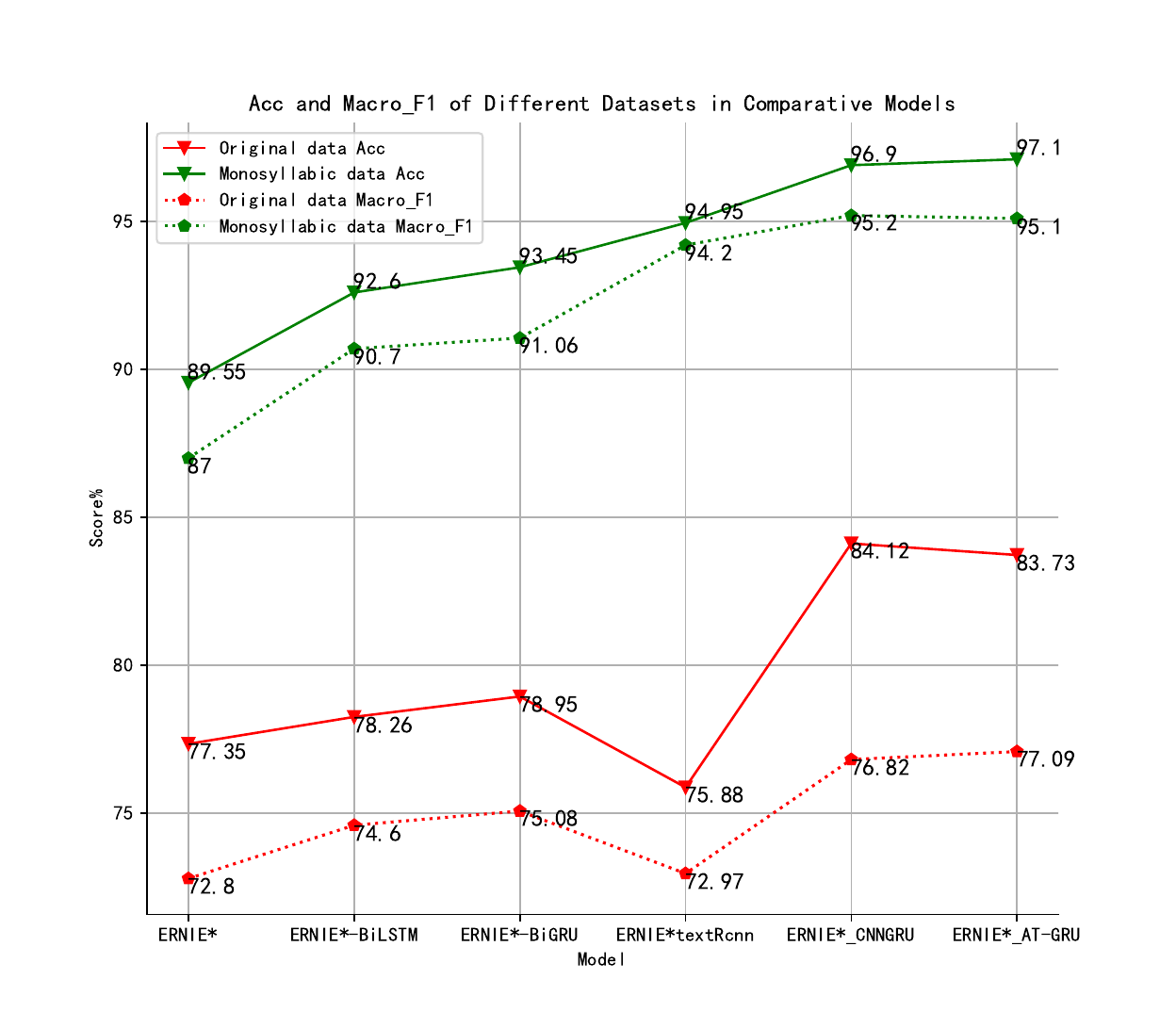}
		\caption{Acc $\&$ Macro$\_$F1 of EP-ERNIE's (ERNIE*) comparative models}
		\label{fig7}
	\end{minipage}
	\begin{minipage}[t]{0.48\textwidth}
		\centering
		\includegraphics[width=8.0cm]{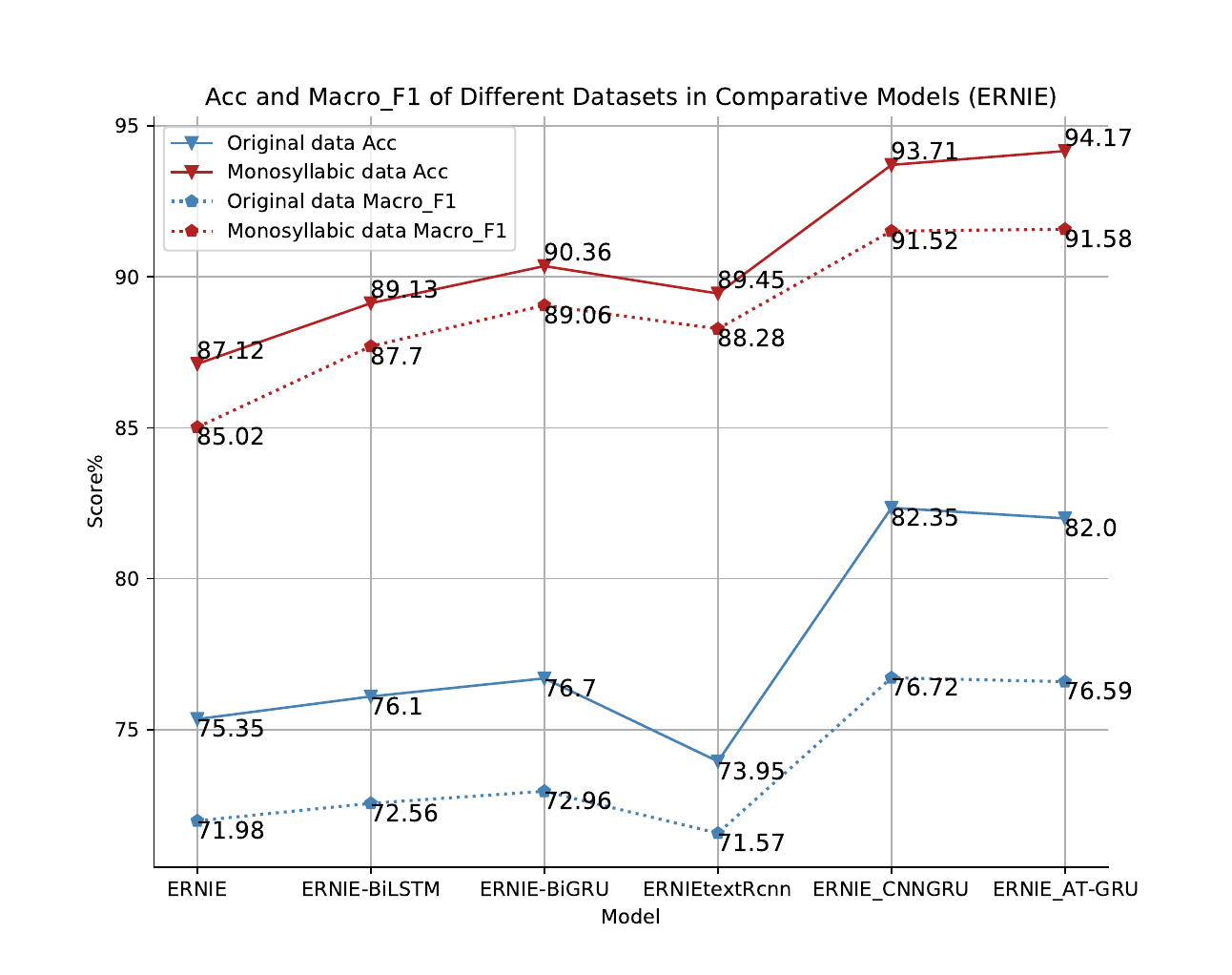}
		\caption{Acc $\&$ Macro$\_$F1 of ERNIE's comparative models (ERNIE)}
		\label{fig8}
	\end{minipage}
\end{figure}	

\begin{figure}[htb]
	\begin{center}
		\includegraphics[width=8.0cm]{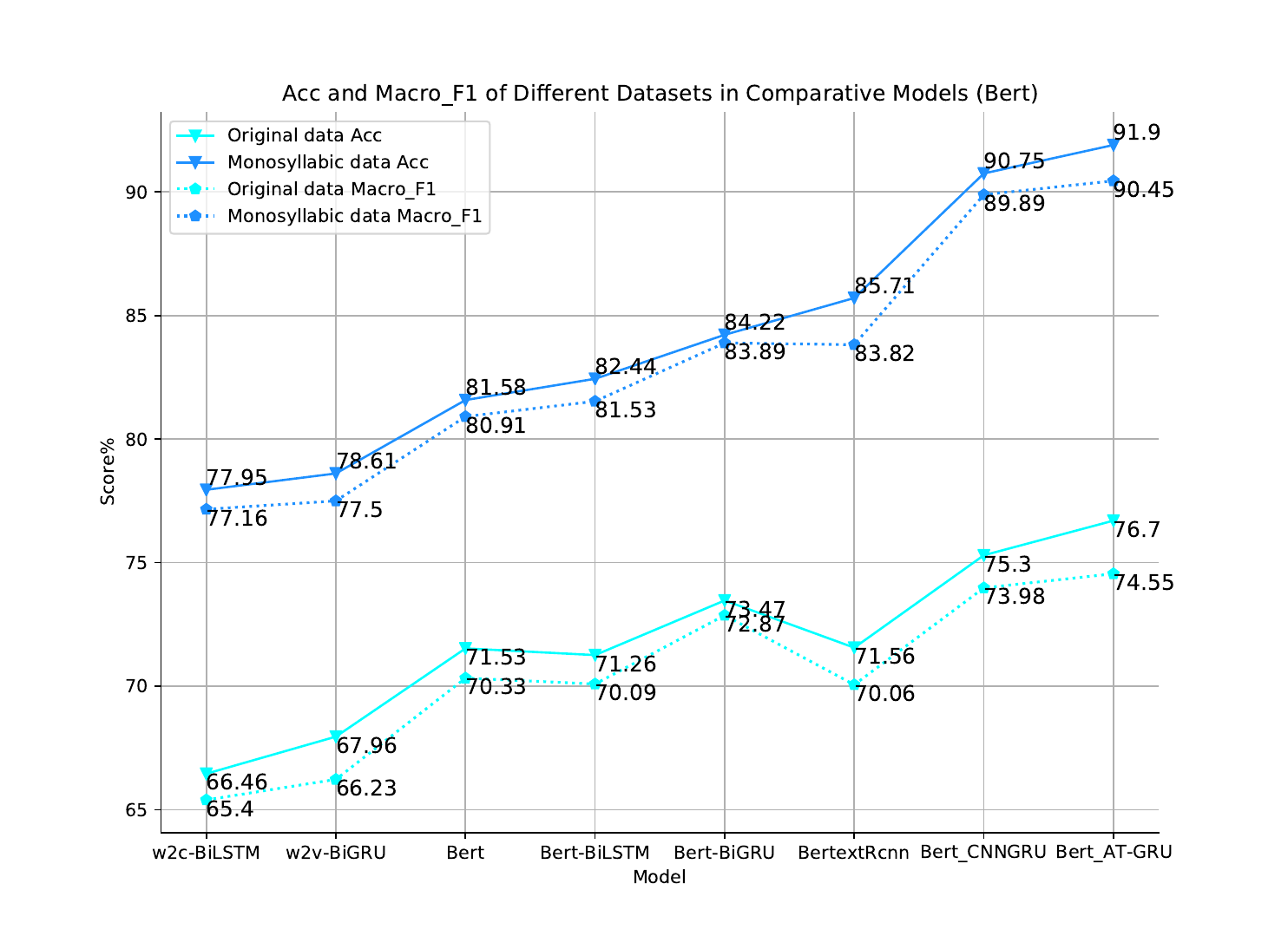}
		\caption{Acc $\&$ Macro$\_$F1 of ERNIE's comparative models (BERT)}
		\label{fig13}
	\end{center}
\end{figure}

\begin{figure}[htb]
	\begin{center}
		\includegraphics[width=8.0cm]{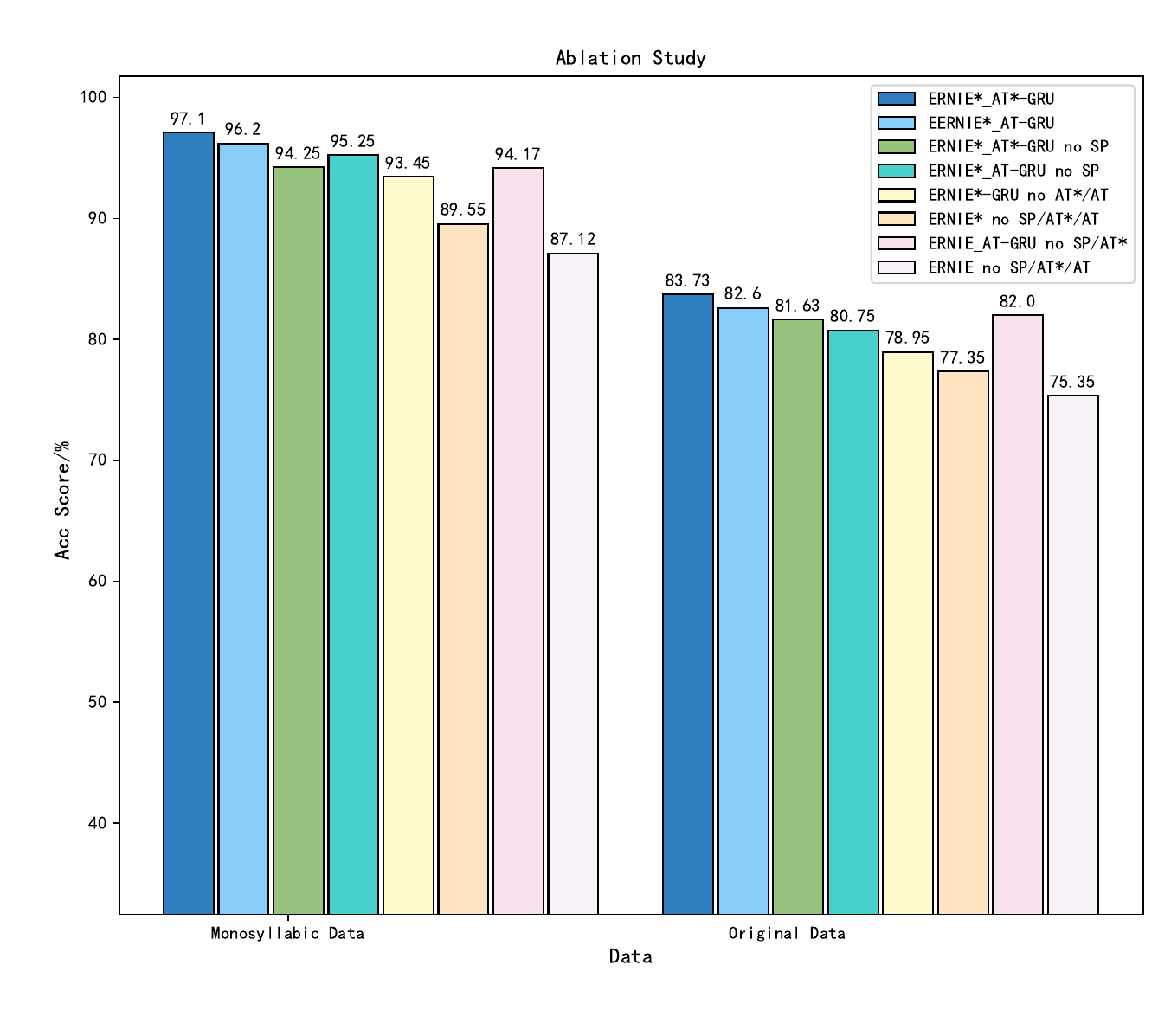}
		\caption{Ablation Study}
		\label{fig9}
	\end{center}
\end{figure}
Acc $\&$ Macro$\_$F1 line chart as Figure \ref{fig7},\ref{fig8},\ref{fig13} confirms that our proposed model in this article outperforms other baseline models in move recognition. The algorithm's performance on monosyllabic data is significantly better than on the original data. Additionally, the improved EP-ERNIE pre-trained model outperforms the original ERNIE pre-trained model in move recognition on both datasets.

Our model integrates complex sentence distinguishing and splitting techniques, EP-ERNIE deep semantic learning and AT-GRU feature recognition. According to the ablation study as Figure \ref{fig9} (EP-ERNIE is marked as ERNIE*, improved self-attention is marked as AT*). The EP-ERNIE structure is necessary to this work, the application of the sentencepiece is helpful in the model, combines with the results table AT-GRU performs well in automatically recognizing unstructured abstracts in Chinese scientific and technological papers. It is evident that the absence of any component significantly impacts the effectiveness of abstract recognition, leading to a reduction in the accuracy of our model.

The experimental results, which compare models within similar classes, demonstrate that the disparities between various pre-trained models and downstream recognition models have an impact on the effectiveness of text semantic learning and feature extraction. Additionally, the complexity of the corpus plays a crucial role in text knowledge acquisition. In this regard, the processing of corpus data and the training of the sentence model using domain-specific texts contribute to the contextual learning of the model. The enhanced model exhibits a substantial improvement in precision and recall compared to other models, indicating its superior performance in text semantic learning and feature extraction. Consequently, it achieves better abstract move recognition result. 

\section{Conclusion and Future Work}
\label{6}
This article proposes an improved EPAG (EP-ERNIE$\_$AT-GRU) framework that incorporates complex sentence distinguishing and splitting, location information and domain-specific training for automatic move recognition of abstracts in unstructured long sentences in Chinese scientific and technological papers. The enhanced EP-ERNIE is utilized for domain-specific deep semantic learning with positional information, AT-GRU is employed for focused feature recognition. Ablation study shows the improved EP-ERNIE and AT-GRU have improved model’s excellent performance so as the evaluation results confirmed. Future research, we will explore the generalization of move recognition models to other types of texts, such as legal, medical, and engineering texts. Additionally, we will investigate the semantic learning of text graph structures to improve  the performance and stability of the model on other agglutinative languages.

\section*{Declaration of competing interest }
The authors declare that they have no known competing financial interests or personal relationships that could have appeared to influence the work reported in this paper.

\bibliographystyle{cas-model2-names}
\bibliography{cas-refs}

\end{document}